\documentclass[conference]{IEEEtran}
\IEEEoverridecommandlockouts
\usepackage{cite}
\usepackage{amsmath,amssymb,amsfonts}
\usepackage{algorithmic}
\usepackage{graphicx}
\usepackage{textcomp}
\usepackage{cleveref}
\usepackage{xcolor}
\def\BibTeX{{\rm B\kern-.05em{\sc i\kern-.025em b}\kern-.08em
    T\kern-.1667em\lower.7ex\hbox{E}\kern-.125emX}}
\begin{document}

\title{
Decentralized Multi-AGV Task Allocation based on Multi-Agent Reinforcement Learning with Information Potential Field Rewards
\thanks{*Corresponding author.}
}

\author{\IEEEauthorblockN{Mengyuan Li, Bin Guo*, Jiangshan Zhang, \\
Jiaqi Liu, Sicong Liu, Zhiwen Yu}
\IEEEauthorblockA{\textit{School of Computer Science} \\
\textit{Northwestern Polytechnical University}\\
Xi’an 710072, China \\
guob@nwpu.edu.cn}

\and
\IEEEauthorblockN{Zhetao Li}
\IEEEauthorblockA{\textit{College of Computer Science} \\
\textit{Xiangtan University}\\
Xiangtan 411105, China \\
liztchina@hotmail.com}
\and
\IEEEauthorblockN{Liyao Xiang}
\IEEEauthorblockA{\textit{John Hopcroft Center for Computer Science} \\
\textit{Shanghai Jiao Tong University}\\
Shanghai 200240, China \\
xiangliyao08@sjtu.edu.cn}

}

\maketitle

\begin{abstract}
Automated Guided Vehicles (AGVs) have been widely used for material handling in flexible shop floors. Each product requires various raw materials to complete the assembly in production process. AGVs are used to realize the automatic handling of raw materials in different locations. Efficient AGVs task allocation strategy can reduce transportation costs and improve distribution efficiency. However, the traditional centralized approaches make high demands on the control center's computing power and real-time capability. In this paper, we present decentralized solutions to achieve flexible and self-organized AGVs task allocation. In particular, we propose two improved multi-agent reinforcement learning algorithms, MADDPG-IPF (Information Potential Field) and BiCNet-IPF, to realize the coordination among AGVs adapting to different scenarios. To address the reward-sparsity issue, we propose a reward shaping strategy based on information potential field, which provides stepwise rewards and implicitly guides the AGVs to different material targets. We conduct experiments under different settings (3 AGVs and 6 AGVs), and the experiment results indicate that, compared with baseline methods, our work obtains up to 47\% task response improvement and 22\% training iterations reduction.
\end{abstract}

\begin{IEEEkeywords}
Multi-agent reinforcement learning, AGVs, decentralized task allocation, information potential field
\end{IEEEkeywords}

\section{Introduction}
Driven by the recent advancements in industry 4.0 and industrial artificial intelligence, the use of autonomous systems in manufacturing enterprises has become inevitable \cite{b1,b2}. Automated Guided Vehicles (AGVs), as a type of flexible intelligent logistics equipment, have a great degree of freedom and play an essential role in flexibly transporting materials and products. AGVs have been hailed as one of the most promising technologies and have been implemented in a variety of shop floors and warehouse logistics operations for material supply \cite{b3,b4}.

\begin{figure}[htbp]
\centerline{\includegraphics[width=0.44\textwidth]{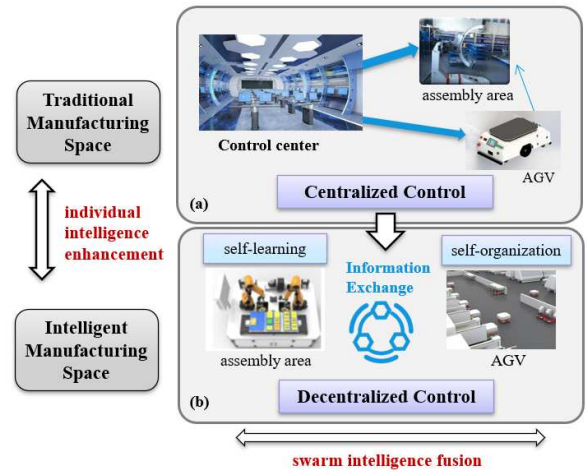}}
\caption{ Centralized control methods and decentralized control methods of AGVs.}
\label{fig1}
\end{figure}

The multi-variety, small-batch, and customized production mode results in more logistics tasks and higher real-time demands. Using AGVs for cooperative transportation can significantly improve efficiency and cut expenses. How to make multiple AGVs collaborate to perform material transportation tasks remains a significant topic in intelligent storage systems \cite{b5,b6}. The traditional approaches are mostly \textit{centralized} control methods (Fig.\ref{fig1} (a)) and consider task assignment as a path planning problem for single or multiple robots \cite{b7,b8}. On one hand, it places extremely high demands on the control center’s computing power and real-time capability. On the other hand, the complexity and dynamic obstacles of the environment can impair the system’s stability and scalability. In comparison to centralized solutions, agent-level \textit{decentralized} task allocation strategies (Fig.\ref{fig1} (b)) evenly distribute computing load and make advantage of agents’ autonomous decision-making ability. 

With the continuous development of Multi-Agent Reinforcement Learning (MARL) \cite{b9}, Reinforcement Learning (RL) has developed the capabilities of autonomous learning and distributed computing. Agents generate their own behaviors, modify their own state information, and accomplish the goal efficiently through cooperation with others \cite{b10}. For example, Lowe et al. \cite{b11} propose the Multi-Agent Deep Deterministic Policy Gradient (MADDPG), which extends the DDPG method to MARL by observing the opponent’s behavior. Meanwhile, a global critic function is constructed to evaluate global state action. The Alibaba team proposes the Bidirectionally Coordinated Network (BiCNet) algorithm \cite{b12} in the pysc2 multi-agent scenario \cite{b13}. Using Bidirectional Recurrent Neural Networks (BRNN) \cite{b14} for implicit communication, BiCNet has demonstrated superior performance in complicated environments. 

However, existing MARL approaches have a number of drawbacks that make them unsuitable for decentralized multi-AGV task allocation directly, such as environmental non-stationarity and partial observability. Additionally, the reward mechanism in multi-agent system is more sophisticated than it is in single-agent system, and the reward-sparsity issue frequently makes training progress difficult to converge. A critical question is how to design an effective reward mechanism that will boost performance and expedite convergence. Information Potential Field (IPF) \cite{b15} is often utilized to tackle the path planning problem. Using the virtual information gradient diffusion of the target position data, the robot can advance to the target position along a specific gradient direction. By including IPF into reward function, the agents' status can be assessed more comprehensively, guiding the agents toward the target positions.

To solve the above challenges, this paper proposes a novel multi-agent reinforcement learning algorithm based on information potential field rewards. We model the decentralized multi AGV task allocation as a Partially Observable Markov Decision Process (POMDP). To address reward-sparsity issue, we propose a reward shaping mechanism based on IPF that provides AGV collaboration with stepwise and implicit direction. Additionally, we apply IPF to the state-of-the-art MADDPG and BiCNet algorithms to prove the superiority of this mechanism. Extensive experiments demonstrate that our methodology can result in considerable performance and convergence improvements. The main contributions of this work are summarized as follows.

(1) The traditional centralized task allocation methods place extraordinarily high demands on control center’s computing power and real-time capability. We innovatively formulate the decentralized multi-AGV task allocation problem as a partially observable Markov decision process, and propose two improved multi-agent reinforcement learning algorithms to achieve coordination among AGVs adapting to different scenarios.

(2) We introduce information potential field to address the reward sparsity issue in decentralized multi-AGV task allocation. It can provide implicit direction for autonomous decision-making and improve the AGV system's cooperation.

(3) We conduct experiments under different settings, and the experiment results show that our strategy obtains up to 47\% task response improvement compared with baseline methods. Additionally, we demonstrate the cooperation mechanism of MADDPG-IPF and BiCNet-IPF. The agents establish a differential preference for each target in MADDPG, while the agents prefer the closest target in BiCNet.

The paper is organized as follows: In Section \uppercase\expandafter{\romannumeral2}, we discuss related literature on collaborative task allocation and multi-agent reinforcement learning. Section \uppercase\expandafter{\romannumeral3} formulates the task allocation problem. In Section \uppercase\expandafter{\romannumeral4}, we model the task allocation problem as a partially observable Markov decision process, and the proposed algorithm is demonstrated. The effectiveness of the method is verified by experiments in section \uppercase\expandafter{\romannumeral5}. Section \uppercase\expandafter{\romannumeral6} gives the conclusions of this study and envisages some future work.

\section{Related Works}
\subsection{Multi-AGV Task Allocation}

Multi-AGV task allocation is a critical part of AGV control, as it seeks to determine the appropriate transit time and equipment for each task. The traditional AGVs task allocation approach is to apply classical optimization algorithms to the production scheduling field, such as genetic algorithm, particle swarm algorithm, ant colony algorithm. Wang et al. \cite{b16} optimize the path selection problem using an improved micro-genetic algorithm that takes into account running time, stopping time, and turning time. Zhang et al. \cite{b17} employ the makespan of jobs as the goal function and the machine and AGV utilization ratios as the comprehensive evaluation function. An improved particle swarm optimization algorithm is developed to solve a reasonable scheduling scheme. Liu et al. \cite{b18} develop a multi-objective mathematical model and integrate with two adaptive genetic algorithms to optimize the task scheduling of AGVs while taking into account the charging task and the AGV's variable speed. Saidi et al. \cite{b19} address the conflict-free AGV path planning problem for job shop scheduling and solve it using a two-stage ant colony algorithm. These algorithms require knowledge of the global environment in order to calculate the optimal policies, and the decision-making capability of a single agent is insufficient in real-world scenarios. The multi-agent system can complete not only a single agent’s goal, but also exceed the efficiency of the single agent, which means that many agents can increase its strength.

\subsection{Multi-Agent Reinforcement Learning}

In multi-agent system, traditional independent Q-learning \cite{b20} or DQN based on experience replay \cite{b21} cannot be applied to a multi-agent environment directly. Because the experience pool's samples become old when the environment changes, the method produced from outdated sample training is frequently not ideal. Therefore, Foerster et al. \cite{b22} propose two strategies for maintaining the DQN experience replay pool's stability. The central idea is to augment the experience buffer with additional information and to undertake importance sampling in order to mitigate the influence of unstable surroundings on multi-agent training. Lowe et al. \cite{b11} propose MADDPG to train a centralized critic for each agent using all agents' policies during training in order to reduce variance by eliminating the non-stationarity. The actor only has local information and the experience buffer records the experiences of all agents. Foerster et al. \cite{b23} propose an actor-critic counterfactual multi-agent (COMA) policy gradient method. COMA is intended for use in both the fully centralized and multiagent credit assignment problems. By comparing the current Q value to the counterfactual, an advantage function can be constructed. In contrast to previous approaches, in Bidirectionally Coordinated Network (BiCNet) \cite{b12}, communication takes place in the latent space, and it also uses parameter sharing. Note that in BiCNet, agents do not explicitly share a message, it might be considered a method for learning cooperation.

Multi-agent reinforcement learning technology provides new ideas for implementing autonomous decision-making of multiple AGVs. Our proposed method utilizes the powerful data representation and decision-making capabilities of deep reinforcement learning to enable self-organizing task assignment of multi-AGV systems.

\subsection{Information Potential Field}

Information Potential Field (IPF) is an effective path planning method. The robot can accomplish the global objective by employing a greedy strategy based on the information gradient. Liu et al. \cite{b15} propose two effective algorithms for constructing IPF: the hierarchical skeleton-based construction algorithm and the value estimation replacement algorithm, both of which achieve a trade-off between energy consumption and convergence speed. Wei et al. \cite{b24} propose efficient parking navigation via a continuous information ascent method. In the first step, a partial differential equation is used to establish a global potential field. In the second step, a Poisson equation is employed to construct the local potential field in the navigation process. Lin et al. \cite{b25} propose an artificial information gradient that is robust and has no local extrema. They use a harmonic function to establish IPF, representing the diffusion of a specific type of event of interest (EoI). Wei et al. \cite{b26} offer a novel heat diffusion equation to efficiently and quickly complete the navigation procedure. The strategy assures that a local information field is sufficiently large to encompass many appropriate targets, and that competition conflicts can be addressed concurrently. The majority of current research directly addresses the path planning problem using the information potential field method. In this paper, the information potential field is utilized to design the reward function of multi-agent reinforcement learning. The reward is evaluated in relation to the information potential value of the AGV location to implicitly steer the AGV to the target position.

\section{Problem Formulation and System Overview}
\subsection{Problem Formulation}
In this section, we will formally define the multi-AGV collaborative task allocation problem. In the manufacturing workshop, processing products typically require various raw materials, which are stored in different locations across the warehouse. AGVs must travel to multiple destinations in order to coordinate transportation tasks. We define the logistic network using $G=(T,V,L)$, where $T$, $V$ and $L$ denote the set of material targets, vehicles and trajectories, respectively. More specifically,

Target set $T$: Each cooperative transportation task entails the movement of $N$ different materials. The material targets $T_i\in T (1\leq i\leq N )$ are randomly dispersed in different places, and the position of the target $T_i$ is represented by $(x_i^T,y_i^T)$.

Vehicle set $V$: we assume that all of $N$ AGVs are modeled as discs with the same radius $D$, i.e., all AGVs are homogeneous. At each timestep $t$, utilize the vector $G_i=\{p_i^t,v_i^t,r_i\}$ to describe the state of the AGV $i$ $(1\leq i\leq N)$, including its position $p_i^t=(x,y)$, velocity $v_i^t=(v_x,v_y)$, and sensing distance $r_i$. The AGV $i$ obtains an observation $o_i^t$ within the sensing range $r_i$, and then compute the action command $a_i^t$ according to the policy $\pi_\theta$, where $\theta$ denotes the policy parameters. The calculated action $a_i^t$ is a velocity $v_i^t$ that directs the AGV toward the task target while avoiding collisions with other robots.

Trajectory set $L$: To wrap up the preceding formulation, we define $L=\{ l_i,i=1,…,N \}$ as the set of trajectories of all AGVs, which are subject to the AGV’s kinematic constraints, i.e.:

\begin{equation}
\begin{aligned}
       v_i^t \backsim \pi_\theta & (a_i^t\mid o_i^t )                   \\
            \Vert v_i^t \Vert \leq & v_i^{max}                      \\
         p_i^t = p_i^{t-1} &+\Delta t \cdot v_i^t                   \\
\forall j \in [1,N],j \neq i,&\left \| p_i^t-p_j^t \right \|>2D
\end{aligned}
\end{equation}

To find an optimal policy, we set an objective by minimizing the expectation of the mean arrival time of all AGVs in the same scenario, which is defined as:
\begin{equation}
{argmin}_{\pi_\theta} E\left [ \frac{1}{N} \sum_{i=1}^{N} t_i  \vert \pi_\theta  \right ] 
\end{equation}

Where $t_i$ is the travel time of the trajectory $l_i$ in $L$ controlled by policy $\pi_\theta$.

Decentralized multi-AGV task allocation can be viewed as a special mobile robot moving path planning problem. AGV decides its target and plans a collision-free course based on its surroundings cognition.

\subsection{System Architecture}
\begin{figure}[htbp]
\centerline{\includegraphics[width=0.45\textwidth]{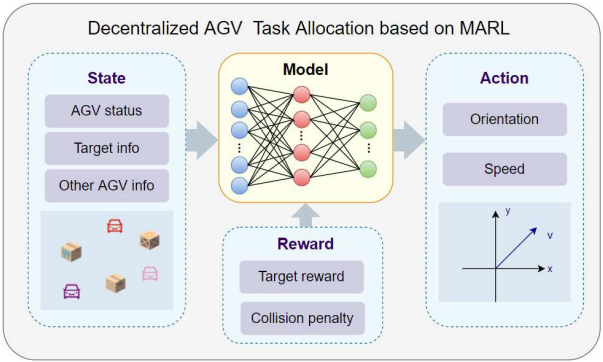}}
\caption{ Architecture of AGVs task allocation approach.}
\label{fig2}
\end{figure}

We propose improved multi-agent reinforcement learning algorithms to solve this problem, the architecture of which is shown as Fig.\ref{fig2}. In real world situations, agents make noisy observations of the true environment state to inform their action selection, typically modeled as a POMDP. Formally, a POMDP can be described as a tuple: $M=(N,S,A,P,R,O)$, where $N$ denotes the number of agents, $S$ represents the system state space, $A$ represents the joint action space of all agents, $P$ is the transition probability function, $R$ is the reward function, and $O$ is the observation probability distribution given the system state $(o\backsim O(s))$. Specific to the problem scenario of AGV collaborative task allocation, the state space $S$ and action space $A$ are specifically designed as follows:

State space $S$: For the AGV task assignment problem, the selection of the state space should not only characterize the attributes of the agents and targets, but also not bring too much computational burden. Therefore, we set the state space as $\{v,p,D_A,D_B\} $, where $\{v,p\}$ is the speed and position of the agent itself, and $ \{D_A,D_B\}$ is the relative distance from the targets and other agents.

Action space $A$: We set the AGV's action space as a one-dimensional vector $\{x,y\}$, the value is $(-1, 1)$, representing the acceleration in the left and right directions and the front and back directions. Combined with the weight and damping of the AGV itself, the velocity of the AGV is computed.

Reward $R$: Our objective is each AGV avoids collisions and self-organizes to different targets as quickly as possible. A reward function is designed to guide a team of AGVs to achieve this objective. we design a target reward when reaching the target position and a collision penalty when a collision occurs. 

When the new tasks arrive, state information is input to the network to determine the action. Following that, the chosen action will be used to route the AGVs to various task targets. The reward function is used to direct model training in this process, allowing the model to learn the ideal strategy.

\section{Methods}

\subsection{Reward Shaping with IPF}
A well-designed reward function can enhance robustness and promote agent collaboration. In the previous section, we discuss a general AGV task allocation framework. In this section, we propose a reward shaping strategy based on information potential field to address the issue of reward sparsity.

\begin{figure}[htbp]
\centerline{\includegraphics[width=0.30\textwidth]{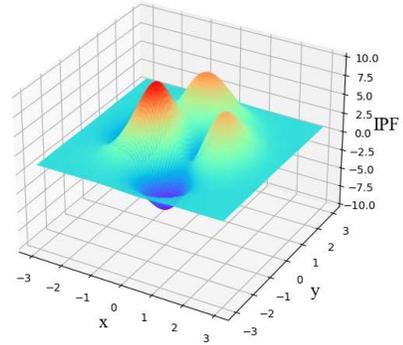}}
\caption{Information Potential Field.}
\label{fig3}
\end{figure}

Information Potential Field (IPF) is introduced to design the reward function $r_{IPF}$, as shown in Fig.\ref{fig3}. We partition the scenario into a bounded grid map, assign a positive information potential value for the location of the target target, and assign a negative information potential value for the location of other AGVs, which can implicitly guide the AGVs to different targets. The targets are set to a maximum potential value of 5, while the other AGV’s positions are set to a minimum potential value of -3. Additionally, we set the information value of some other nodes to 0, often nodes on the network boundary, in order to enforce a gradient throughout the network. The remaining nodes compute the information potential field using Jacobi iterations. Each non-boundary node iterates:
\begin{equation}
\Phi^{k+1}(u) \leftarrow \frac{1}{d(u)} \sum _{v\in N(u)} \Phi^{k}(u)
\end{equation}

\begin{figure}[htbp]
\centerline{\includegraphics[width=0.30\textwidth]{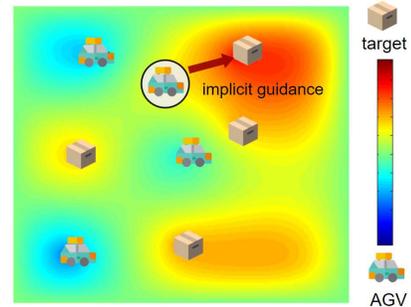}}
\caption{IPF provides implicit guidance for agent’s decision-making.}
\label{fig4}
\end{figure}

Where $\Phi^k (u)$ is the value of node $u$ in the $k-th$ iteration. $N(u)$ signifies the set of $u$’s neighbors, while $d(u)$ denotes the degree of $u$. Each position will have a corresponding information potential value after iteration. The AGV obtains the reward value $r_{IPF}$ according to the information potential value of the position at the time step $t$. As illustrated in Fig.\ref{fig4}, the IPF value around the target location is high, and the gravitational range grows more vast when several targets are gathered. When another AGV is already in close proximity to the target, the reward is reduced, essentially avoiding multiple AGVs competing for the same target.

Along with $r_{IPF}$ for implicit guidance, we design a target reward $r_g$ and a collision penalty $r_c$ for explicit guidance. The target reward $r_g$ and the collision penalty $r_c$ are specified as follows:
\begin{equation}
r_g=-\sum_i min_j(d_{ij})
\end{equation}

\begin{equation}
r_c=\left\{
\begin{array}{rcl}
-1       &      & {if \Vert p_i^t-p_j^t\Vert \leq 2R }\\
0        &      & {otherwise}
\end{array} \right.
\end{equation}

Where $d_{ij}$ is the distance between task target $j$ and AGV $i$. Additionally, when the AGV collides with other AGVs in the environment, it incurs a $r_c$ penalty.

In general, we hope that when a new handling task arrives, the AGV system can self-organize and complete it in the shortest time possible. Based on the observed information, AGVs must plan a collision-free path to different material targets. We use the sum of $r_{IPF}$, $r_g$ and $r_c$ to represent the reward $r$ acquired by AGV $i$ at time step $t$, as seen in (\ref{eqs1}), directing the AGV system to achieve self-organizing task assignment. $r_g$ incentivizes the presence of precisely one agent near each target. $r_c$ wishes for the fewest potential collisions. $r_{IPF}$ provides an implicit shove to the AGV, guiding it to the target place in a distributed fashion.
\begin{equation}
\label{eqs1}
r_i^t=(r_{IPF})_i^t+(r_g)_i^t+(r_c)_i^t
\end{equation}

\subsection{The Algorithm Design}
In multi-agent training, we focus on two algorithms based on the actor-critic framework, MADDPG and BiCNet. These two algorithms offer the following advantages over other MADL algorithms. MADDPG does not require explicit communication rules, is applicable to a wide variety of contexts, including cooperative, competitive, and mixed environments, and is capable of solving the non-stationary problem associated with multi-agent environments. All agents in BiCNet share models and parameters and build communication channels in the hidden layer, enabling any number of agents to cooperate. These two algorithms approach issues differently, and there are clear distinctions in the model structure, loss function, and other factors.

\subsubsection{MADDPG-IPF}
\begin{figure}[htbp]
\centerline{\includegraphics[width=0.45\textwidth]{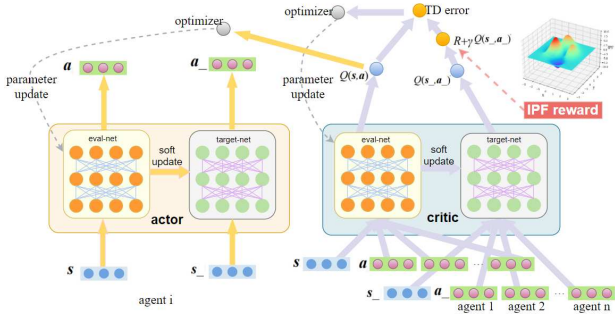}}
\caption{ The structure of MADDPG-IPF.}
\label{fig5}
\end{figure}

MADDPG \cite{b11} adopts centralized training with distributed execution method. Each agent trains a critic network that requires global information and an actor network that only requires local knowledge. The actor chooses the best action for a given state by optimizing the neural network parameters $\theta$. The critic evaluates the action generated by the actor by computing the temporal difference error. The MADDPG algorithm network structure is shown in Fig.\ref{fig5}.

The policy gradient is calculated as:
\begin{equation}
\begin{aligned}
\nabla_{\theta_i} J(\mu _i)=& E_{x,a\backsim D}[\nabla_{\theta_i}\mu_i (a_i \vert o_i)\cdot \\&\nabla_{a_i}Q_i^\mu (x,a_1,\dots,a_n)\vert _{a_i=\mu_i(o_i)}]
\end{aligned}
\end{equation}

Among them, $o_i$ represents the observation of the agent $i$, and $x=[o_1,…,o_n]$ represents the observation vector. $Q^\mu_i (x,a_1,…,a_n )$ represents the centralized state-action function of the  agent $i$. The experience replay buffer $D$ contains $(x,x^,,a_1,\dots,a_n,r_1,\dots,r_n)$ these tuples, which acts as the knowledge base of the agent, storing the experience of all agents. 

The action-value function $Q_i^\mu$ is updated based on:
\begin{equation}
\begin{aligned}
& y=  r_i(s,a)+\lambda Q_i^{\mu^\prime}max_\theta(x^\prime,a_1^\prime,\dots,a_n^\prime)\vert_{a_j^\prime=\mu_j^\prime(o_j)} \\&
L(\theta_i) = E_{x,a,r,x^\prime}[(Q_i^\mu(x,a_1,\dots,a_n)-y)^2]
\end{aligned}
\end{equation}

Among them, $Q_i^{\mu^\prime}$ represents the target network, and $\mu^\prime=[\mu_1^\prime,\mu_2^\prime,\dots,\mu_n^\prime] $is the parameter $\theta_j^\prime$ of the target network that has a lagging update. 

\subsubsection{BiCNet-IPF}

\begin{figure}[htbp]
\centerline{\includegraphics[width=0.45\textwidth]{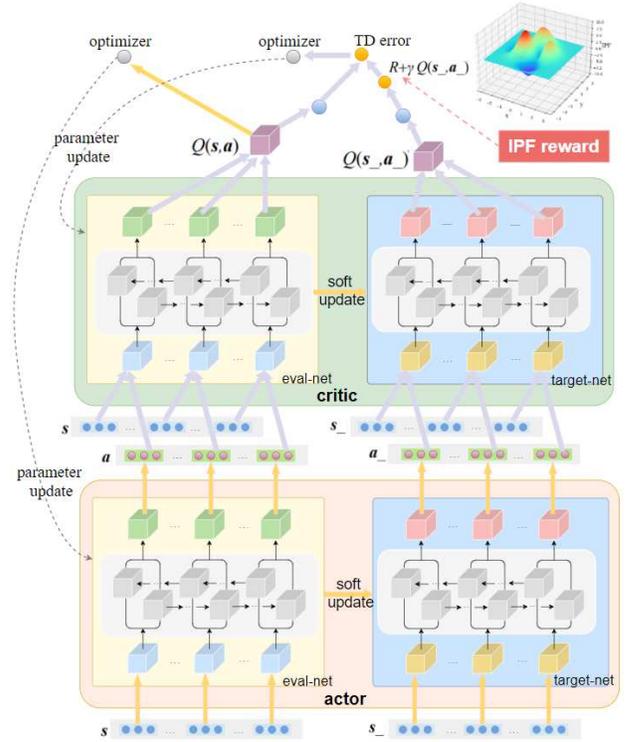}}
\caption{ The structure of BiCNet-IPF.}
\label{fig6}
\end{figure}

BiCNet \cite{b12} is still based on the actor-critic framework, and the network structure as illustrated in Fig.\ref{fig6}. The actor and the critic are both constructed using a bidirectional recurrent neural network. Through implicit communication, the actor shares observation and returns action for each agent. Each agent has the ability to retain its own internal state and communicate with other agents.

We denote the objective of a single agent $i$ by $J_i (\theta)$, that is to maximize its expected cumulative individual reward $r_i$ as $J_i (\theta)=E_{s\backsim \rho_{a_\theta}^\tau } [r_i (s,a_\theta (s))]$. Therefore, we can get the objective of $N$ agents denoted by $J(\theta)$ as follows:

\begin{equation}
J(\theta)=E_{s\backsim \rho_{a_\theta}^\tau } [\sum_{i=1}^{N}r_i (s,a_\theta (s))]
\end{equation}

Combined with the deterministic policy gradient, we have the policy gradient as follows:

\begin{equation}
\nabla_\theta J(\theta)=E_{s\backsim \rho_{a_\theta}^\tau(s)}[\sum_{i=1}^{N}\sum_{j=1}^{N} \nabla_\theta a_{j,\theta} \cdot \nabla_{a_j} Q_i^{a_\theta} (s,a_\theta (s)) ]
\end{equation}

In training the critic network, using the sum of square loss, the gradient can be written as in (\ref{eqsx}), where $\xi$ is the parameter of the Q-network:

\begin{equation}
\label{eqsx}
\begin{aligned}
\nabla_\xi L(\xi)=& E_{s\backsim \rho_{a_\theta}^\tau(s)}[\sum_{i=1}^{N}( r_i (s,a_\theta (s))+\lambda Q_i^\xi (s^\prime,a_\theta (s^\prime)) 
\\ & - Q_i^\xi (s,a_\theta (s)))\cdot \nabla_{\partial\xi} Q_i^{\xi} (s,a_\theta (s))]
\end{aligned}
\end{equation}

In different agents, the parameters are shared, hence the number of parameters is independent of the number of agents. Parameter sharing leads to a compact model that speeds up the learning process.

\section{Evaluation}
\subsection{Experimental Settings}
\begin{figure}[htbp]
\centerline{\includegraphics[width=0.25\textwidth]{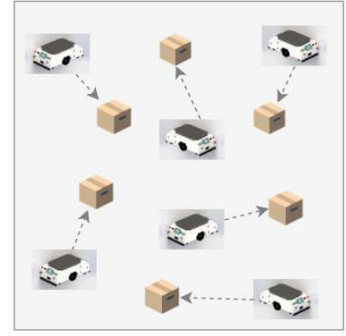}}
\caption{Cooperative task allocation.}
\label{fig7}
\end{figure}

In order to conduct experiments, we build an AGV task allocation simulator based on a multi-agent environment \cite{b11}, which comprises of $N$ AGVs and $N$ tasks inhabiting a two-dimensional world with continuous space and discrete time (see Fig.\ref{fig7}). For MARL algorithms, as the number of agents increases, the joint state-action space increases exponentially, which makes the task intractable. Therefore we verify the robustness of the proposed methods under two scenarios: a 3 AGVs and 3 tasks simple scenario and a 6 AGVs and 6 tasks complex scenario, referred to as 3V3 scenario and 6V6 scenario. In each scenario, the position of the AGV and the position of the task are randomly generated. Taking into account the actual scenario, we define boundaries around the simulator, within which the agent can only move. We hope that the AGV can learn to disperse to different task targets in the shortest time and avoid collisions as much as feasible. Performance is measured by average task response rate, average reward, and average time:

\textit{Average task response rate:} the number of tasks completed by $N$ AGVs in the entire test epochs divided by the total number of tasks generated.

\textit{Average reward:} the rewards obtained by $N$ AGVs at each time step, calculated using the formula $R=-\sum_{i} min_j (d_{ij})-C$.

\textit{Average time:} the total time required for $N$ AGVs to execute all tasks (for example, in the 3V3 scenario, the three AGVs have reached the three task targets correctly).

\subsection{ Performance Comparison}
In this subsection, the performance of following methods is extensively evaluated by the simulation.

\textbf{MADDPG-MiniDist:} The MiniDist is a global reward that sums the distance between each task target and its nearest agent. The shorter the distance between two targets, the larger the reward.

\textbf{MADDPG-Greedy:} The Greedy is an individual reward. When an agent approaches the task target, it receives a positive reward, which rises as the distance between the agent and the task target decreases.

\textbf{MADDPG-IPF:} The IPF as we discussed in Section 4.

Additionally, \textbf{BiCNet-MiniDist}, \textbf{BiCNet-Greedy} and \textbf{BiCNet-IPF} are similar to the above. The Q-network and policy network in MADDPG are parameterized by three fully connected layers. The Q-network and policy network in BiCNet are based on the bi-directional RNN structure. Both the input and output modules are made up of four fully connected layers.

Each model is trained for 30k epochs in the 3V3 scenario. For the 6V6 scenario, the action space and state space dimensions are greatly increased, necessitating the use of additional rounds. As a result, each model based on MADDPG is trained for 50k epochs, and each model based on BiCNet, a more complicated network structure, is trained for 90k epochs. Finally, we execute 300 epochs for testing on each model in the two scenarios, and the results are presented in Table \ref{table1} and Table \ref{table2}.

\begin{table}[]
\caption{model performance in 3V3 scenario}
\label{table1}
\begin{center}
\begin{tabular}{|l|l|l|l|}
\hline
                & \begin{tabular}[c]{@{}l@{}}Average task\\ response rate\end{tabular} & Average reward & Average time \\ \hline
MADDPG-MiniDist & 88.64\%                                                              & -101.1         & 14.3         \\ \hline
MADDPG-Greedy   & 88.67\%                                                              & -116.61        & 11.8         \\ \hline
MADDPG-IPF      & 95.00\%                                                              & -85.8          & 11.1         \\ \hline
BiCNet-MiniDist & 93.03\%                                                              & -71.5          & 10.4         \\ \hline
BiCNet-Greedy   & 73.56\%                                                              & -143.8         & 10.2         \\ \hline
BiCNet-IPF      & 97.58\%                                                              & -65.8          & 9.7          \\ \hline
\end{tabular}
\end{center}
\end{table}

\begin{table}[]
\caption{model performance in 6V6 scenario}
\begin{center}
\label{table2}
\begin{tabular}{|l|l|l|l|}
\hline
                & \begin{tabular}[c]{@{}l@{}}Average task\\ response rate\end{tabular} & Average reward & Average time \\ \hline
MADDPG-MiniDist & 69.22\%                                                              & -438.5         & 17.7         \\ \hline
MADDPG-Greedy   & 46.06\%                                                              & -675.0         & 17.1         \\ \hline
MADDPG-IPF      & 80.22\%                                                              & -371.5         & 16.2         \\ \hline
BiCNet-MiniDist & 80.44\%                                                              & -249.8         & 16.0         \\ \hline
BiCNet-Greedy   & 44.56\%                                                              & -664.5         & 17.5         \\ \hline
BiCNet-IPF      & 91.61\%                                                              & -241.1         & 15.6         \\ \hline
\end{tabular}
\end{center}
\end{table}

MADDPG-IPF achieves a task response rate of 95\% in the 3V3 scenario, an increase of approximately 6\% over the other MADDPG models. Comparing the results of MADDPG-IPF and BiCNet-IPF, the BiCNet-IPF consistently outperforms MADDPG-IPF, possibly because of implicit communication, which enables better decision-making with more information. In the more complex 6V6 scenario, BiCNet-IPF achieves a task response rate of 91.61\%, a significant advantage over all other models. Although MADDPG-IPF is not as good as the best approach, it still achieves an 80.22\% task response rate. In general, the global reward (MiniDist) assigns the same reward to all agents without regard of their contributions, which may encourage slothful agents. In comparison, the local reward (Greedy) only provides different local rewards to each agent based on individual behavior, leading to selfish agents. IPF reward incorporates global and local information and gives ongoing rewards at each step, allowing the agent to improve its performance on various task targets.

\subsection{The Effectiveness of IPF}
\begin{figure}[htbp]
\centerline{\includegraphics[width=0.47\textwidth]{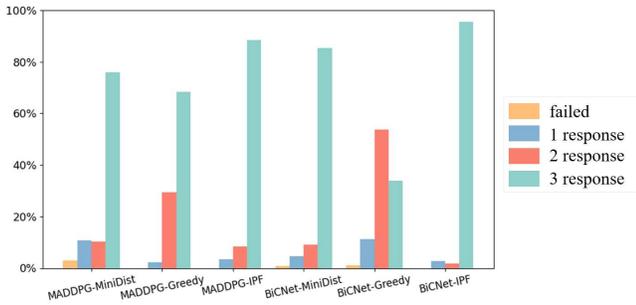}}
\caption{Task response rate per round. For 3 response, all three tasks are completed.}
\label{fig8}
\end{figure}

Along with the performance comparisons mentioned above, we examine the task completion of each round of three AGVs under different reward designs in 3V3 scenario. As shown in the Fig.\ref{fig8}, after applying the IPF reward mechanism, the agents can complete all tasks mostly in a distributed manner. IPF can significantly reduce the likelihood of multiple AGVs competing for the same target by offering implicit guidance. Global rewards may lead to laziness, so the agents inspired by MiniDist sometimes reach the target nearby but stagnate, resulting in worse task response than IPF. The Greedy reward frequently motivates agents to fight for a single task target, resulting in suboptimal performance.

\begin{figure}[htbp]
\centerline{\includegraphics[width=0.47\textwidth]{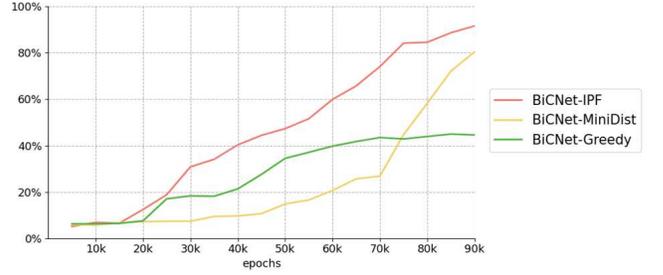}}
\caption{Convergence comparison. Average task response rate under different reward mechanisms during the training phase. }
\label{fig9}
\end{figure}

Convergence is assessed by examining the average task completion rate of BiCNet during the training phase under the challenging 6V6 scenario. As illustrated in Fig.\ref{fig9}, the approach using IPF can achieve a 40\% task response rate after 40k epochs and 60\% task response rate after 60k epochs. Due to the fact that BiCNet-Greedy is an individual reward network, its convergence rate is slower. The agents inspired by MiniDist are unable to acquire vital knowledge in the first 60k epochs, but there performance improves significantly after 70k epochs. 

\subsection{Implicit Cooperation Mechanism Analysis}
\begin{figure}[htbp]
\centerline{\includegraphics[width=0.43\textwidth]{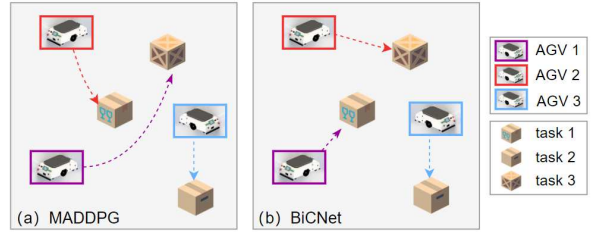}}
\caption{The cooperation mechanism of MADDPG and BiCNet. In (a), the agents develop a differential preference for each task, e.g. the 1-th agent prefers task 1. In (b), the agents tend to complete the nearest task.}
\label{fig10}
\end{figure}

In the 6V6 scenario, by numbering each AGV and each task, we observe an interesting phenomenon: the 2-th AGV and 6-th AGV directed by MADDPG always arrive at the identical 3-th task, resulting in no AGV reaching the 1-th task. However, this will not occur in BiCNet. Thus, we count the task targets achieved by each agent of MADDPG-IPF and BiCNet-IPF in the 3V3 scenario and 6V6 scenario, and investigate the cooperation mechanism of the two methods MADDPG and BiCNet, as shown in Fig. \ref{fig10}.

\begin{figure}[htbp]
\centerline{\includegraphics[width=0.47\textwidth]{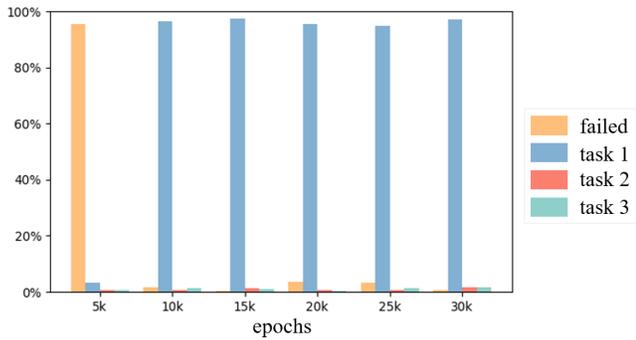}}
\caption{ The 1-th agent’s preference in MADDPG. After fully training the model, the 1-th agent tends to complete task 1, but rarely chooses task 2 and task 3.}
\label{fig11}
\end{figure}

\begin{figure}[htbp]
\centerline{\includegraphics[width=0.47\textwidth]{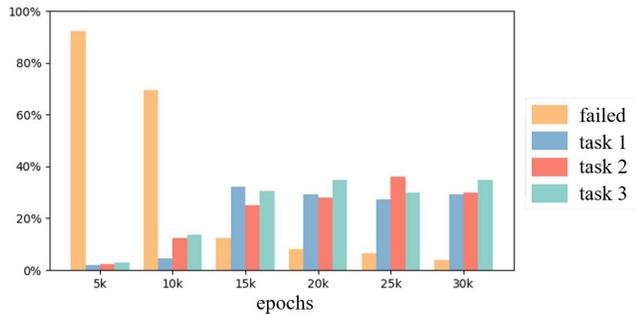}}
\caption{The 1-th agent’s preference in BiCNet. After fully training the model, the probabilities of the 1-th agent choosing three tasks are similar.}
\label{fig12}
\end{figure}

We discover that what MADDPG learned is each agent's preference for a certain fixed task. As illustrated in Fig.\ref{fig11}, while training 30k epochs in the 3V3 scenario, 97\% of the epochs of 1-th AGV chooses the 1-th task. What BiCNet learned is the choice of each agent for the closest task targets. As shown in Fig.\ref{fig12}, the reach rate of 1-th AGV for the three tasks in 30k epochs is approximately 30\%, and it does not show exceptional performance for a particular task. 

In terms of this phenomenon, we argue that under MADDPG framework, each agent has an independent network structure and takes decisions based on local observations. Therefore, by continuously strengthening the rewards obtained at a particular task target during initial training, the agent will prefer it. While all agents in BiCNet share parameters and communicate implicitly via the bi-directional RNN, each agent coordinates with others and moves toward the nearest task target.

\section{Conclusion}
In this paper, we first formulated the AGVs task allocation problem in logistics networks as a partially observable Markov decision process. Given this setting, we introduced the information potential field optimization reward mechanism and proposed two cooperative multi-agent reinforcement learning algorithms to solve the problem. Extensive experiments demonstrate that our new approach can stimulate cooperation among agents and give rise to a significant improvement in both performance and convergence. For future work, we will create more multi-agent coordination and communications scenarios considering complex operation situations and uncertainties. Another interesting and practical direction to develop is to use a heterogeneous agent setting with individual specific feature to improve collaboration. 

\section*{Acknowledgment}

This work was partially supported by the National Science Fund for Distinguished Young Scholars(62025205), National Key R\&D Program of China(2019YFB1703901), and the National Natural Science Foundation of China (No. 62032020, 61960206008, 61725205).

%\section*{References}

\end{document}